\documentclass[acmtog]{acmart}
\usepackage{lipsum}
\usepackage{xcolor}
\usepackage{bbding}

\AtBeginDocument{%
  \providecommand\BibTeX{{%
    Bib\TeX}}}

\setcopyright{rightsretained}
\acmJournal{TOG}
\acmYear{2024} \acmVolume{43} \acmNumber{4} \acmArticle{72} \acmMonth{7}\acmDOI{10.1145/3658147}

\begin{document}

\def\name{DressCode: Autoregressively Sewing and Generating Garments from Text Guidance}

\title{\name}


\author{Kai He}
\email{hekai@shanghaitech.edu.cn}
\orcid{0000-0002-8384-094X}
\affiliation{%
  \institution{ShanghaiTech University and Deemos Technology Co., Ltd.}
  \city{Shanghai}
  \country{China}
}

\author{Kaixin Yao}
\email{yaokx2023@shanghaitech.edu.cn}
\orcid{0009-0005-2056-6057}
\affiliation{%
  \institution{ShanghaiTech University and NeuDim Technology Co., Ltd.}
  \city{Shanghai}
  \country{China}
}

\author{Qixuan Zhang}
\email{zhangqx1@shanghaitech.edu.cn}
\orcid{0000-0002-4837-7152}
\affiliation{%
  \institution{ShanghaiTech University and Deemos Technology Co., Ltd.}
  \city{Shanghai}
  \country{China}
}

\author{Jingyi Yu}\authornote{Corresponding author.}
\email{yujingyi@shanghaitech.edu.cn}
\orcid{0000-0001-9198-6853}
\affiliation{%
  \institution{ShanghaiTech University}
  \city{Shanghai}
  \country{China}
}

\author{Lingjie Liu}\authornotemark[1]
\email{lingjie.liu@seas.upenn.edu}
\orcid{0000-0003-4301-1474}
\affiliation{%
  \institution{University of Pennsylvania}
  \city{Philadelphia}
  \country{USA}
}

\author{Lan Xu}\authornotemark[1]
\email{xulan1@shanghaitech.edu.cn}
\orcid{0000-0002-8807-7787}
\affiliation{%
  \institution{ShanghaiTech University}
  \city{Shanghai}
  \country{China}
}


\begin{abstract}
\label{sec:abstract}
    Apparel's significant role in human appearance underscores the importance of garment digitalization for digital human creation. Recent advances in 3D content creation are pivotal for digital human creation. Nonetheless, garment generation from text guidance is still nascent. We introduce a text-driven 3D garment generation framework, \textit{DressCode}, which aims to democratize design for novices and offer immense potential in fashion design, virtual try-on, and digital human creation. We first introduce SewingGPT, a GPT-based architecture integrating cross-attention with text-conditioned embedding to generate sewing patterns with text guidance. We then tailor a pre-trained Stable Diffusion to generate tile-based Physically-based Rendering (PBR) textures for the garments. 
By leveraging a large language model, our framework generates CG-friendly garments through natural language interaction. It also facilitates pattern completion and texture editing, streamlining the design process through user-friendly interaction. This framework fosters innovation by allowing creators to freely experiment with designs and incorporate unique elements into their work. With comprehensive evaluations and comparisons with other state-of-the-art methods, our method showcases superior quality and alignment with input prompts. User studies further validate our high-quality rendering results, highlighting its practical utility and potential in production settings. Our project page is \textcolor{magenta}{\href{https://IHe-KaiI.github.io/DressCode/}{https://IHe-KaiI.github.io/DressCode/}}.

\end{abstract}


\begin{teaserfigure}
    \centering
  \includegraphics[width=\textwidth]{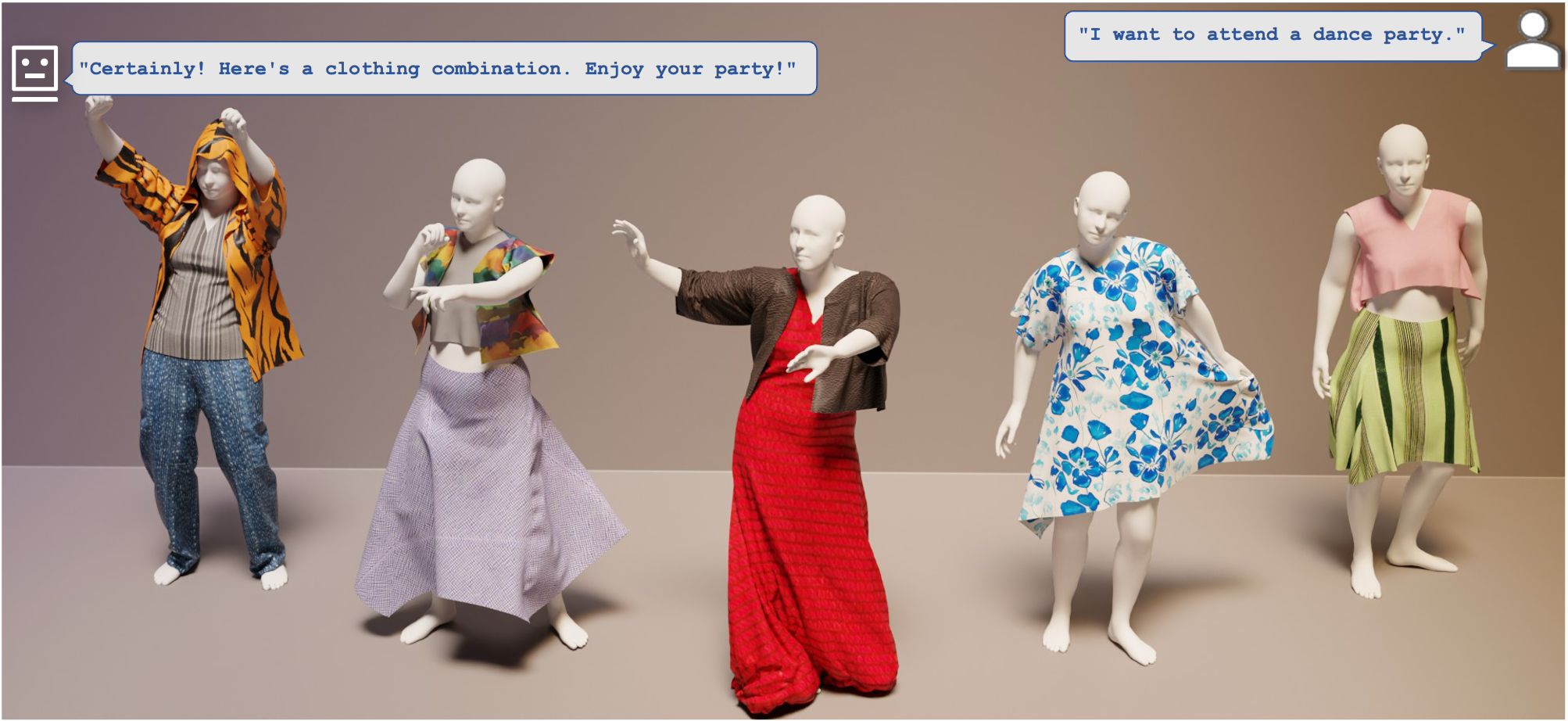}
  \caption{Our \textit{DressCode} generates CG-friendly customized garments with sewing patterns and PBR textures under natural text guidance, enabling post-editing, animation, and high-quality rendering.}
  \label{fig:teaser}
\end{teaserfigure}


\maketitle

\section{Introduction}
\label{sec:intro}
    
Apparel substantially influences the appearance of us humans, and hence garment digitalization has emerged as a vital component of digital human creation. An effective digital garment creation tool should enable users to customize garments to depict the individuality and diversity that make up our physical world, with various garment traits like sewing patterns, styles, or materials. The creation process also needs to match specific themes and be convenient, as simple as chatting with AI agents like ChatGPT.

Recent years have witnessed tremendous progress in text-driven asset generation, triggered by the large-scale language models~\cite{achiam2023gpt, radford2021learning}. It democratizes the accessible text-driven creation of diverse assets for novices, including images~\cite{rombach2022high}, generic 3D objects~\cite{liu2023zero, poole2022dreamfusion}, human hair~\cite{zhou2023groomgen}, face or body~\cite{liao2023tada, zhang2023dreamface}. What is still missing is the garment. Moreover, naively applying avatar or general generation~\cite{liao2023tada, poole2022dreamfusion} for the garment category is suboptimal since they turn to generate mesh or neural fields that are incompatible with digital garment production workflow. 

In contrast, for our graphics community, the dominant representation of garments is sewing patterns, which facilitates both physical simulation and animation in a CG-friendly fashion~\cite{blender2022, maya}.
For sewing pattern generation, early methods~\cite{umetani2011sensitive, berthouzoz2013parsing} only use simple partial modules in the workflow like parsing or draping to 3D. With the consolidation of more advanced datasets~\cite{KorostelevaGarmentData}, recent work enables sewing pattern generation from point clouds~\cite{korosteleva2022neuraltailor} or images~\cite{liu2023sewformer}. 
However, they largely overlook the generation through more natural language interactions, let alone handling the vivid generation with desired texture patterns or physically-based materials, which could significantly speed up the preliminary stage of garment design. 
Furthermore, through the natural interactions of text prompts, novices without professional skills in complex design software can directly describe and transform their ideas into creations. 
This significantly lowers the design barriers, allowing more newcomers to participate in the creative process. Most importantly, generative models introduce diversity to designed text prompts, generate varied types of garments conforming to the prompts, and hence stimulate designer creativity.

In this paper, we propose \textit{DressCode}, a 3D garment generation framework that generates high-quality garments via natural language interaction. As illustrated in Figure~\ref{fig:teaser}, DressCode allows users to customize garments with preferred sewing patterns and physically-based texture details through text interaction. The resulting garments can be seamlessly integrated with CG pipelines, supporting post-editing and animation while ensuring high-quality rendering.

Notably, the garments' highly symmetric and structured nature, with uniform panels and stitching, leads to convenient conversion from sewing patterns to discrete ``codes.''
To this end, inspired by powerful language generation apt for this nature, we introduce SewingGPT, a GPT-based architecture for sewing pattern generation. Specifically,  we adopt a novel quantization process to translate the sewing patterns into token sequences and subsequently utilize a decoder-only Transformer with text-conditioned embedding for token prediction. We utilize the pre-trained CLIP \cite{radford2021learning} model to encode prompts as conditional embeddings, benefiting from CLIP's generalized capability in multimodal understanding. For effective training, we apply GPT-4V~\cite{achiam2023gpt} on the existing dataset~\cite{KorostelevaGarmentData} to detail diverse garment types and shapes with rich text prompts. Once trained, our SewingGPT autoregressively generates quantized sewing patterns with efficient text interactions, which have been unseen before.

To achieve high-quality garment rendering, we progressively tailor a pre-trained Stable Diffusion model \cite{rombach2022high} to generate tile-based Physically-based Rendering (PBR) textures from text prompts. We first fine-tune the U-Net architecture~\cite{ronneberger2015u} of the diffusion model within latent space to generate the diffuse attribute, and then fine-tune the various VAE~\cite{kingma2013auto} decoders to generate normal and roughness maps separately.
We showcase the capability of DressCode to generate CG-friendly garments with rich sewing patterns and PBR textures from text prompts. We also demonstrate the versatility of our approach, including a ChatGPT-like conversational agent for interactive garment generation, garment completion from partial inputs, and user-friendly texture editing.
To summarize, our main contributions include:

\begin{itemize}
    \item We propose a first text-driven garment generation pipeline with high-quality garment sewing patterns and physically-based textures.
    
    \item We introduce a novel generative paradigm for sewing patterns as a sequence of tokens, achieving high-quality autoregressive generation via text guidance.

    \item We tailor a diffusion model for vivid texture generation of garments from text prompts and showcase interaction-friendly applications for garment generation, completion, and editing.

\end{itemize}

\section{Related Work}
\label{sec:related}
    
\paragraph{Garment Sewing Pattern Modeling}

Sewing pattern representation is vital for garment modeling. Recent studies have delved into sewing pattern reconstruction \cite{pietroni2022computational, jeong2015garment, yang2018physics, wang2018learning, sharp2018variational, su2022deepcloth}, generation \cite{shen2020gan}, draping \cite{de2023drapenet, li2023isp, berthouzoz2013parsing} and editing \cite{bartle2016physics, umetani2011sensitive, qi2023personaltailor}. Early research \cite{chen2015garment} employs a search in a pre-defined database of 3D garment parts for garment reconstruction. Studies \cite{jeong2015garment, su2020mulaycap, yang2018physics} use parametric sewing patterns, optimizing them for garment reconstruction from images. \cite{wang2018learning} advances this by applying deep learning to discern a shape space for sewing patterns and various input modalities. \cite{bang2021estimating, sharp2018variational} utilize surface flattening for sewing pattern reconstruction from 3D human models.

Recently, some work \cite{zhu2020deep, goto2021data, korosteleva2022neuraltailor} adopt data-driven approaches for reconstruction. \cite{goto2021data} utilizes a deep network with surface flattening for sewing pattern reconstruction from 3D geometries. \cite{KorostelevaGarmentData} generates a sewing pattern dataset covering a wide range of garment shapes and topologies. NeuralTailor \cite{korosteleva2022neuraltailor} offers advanced sewing pattern reconstruction using a hybrid network to predict garment panels and stitching information from point cloud input. \cite{chen2022structure} introduces a CNN-based model capable of predicting garment panels from single images, using PCA to simplify the panel data structure. \cite{liu2023sewformer} creates a comprehensive dataset featuring diverse garment styles and human poses and introduces a two-level Transformer network, achieving state-of-the-art sewing pattern reconstruction from single images. \cite{korosteleva2023garmentcode} designs the first DSL for garment modeling, enabling users to do rich garment designs using interchangeable, parameterized components. \cite{li2023diffavatar} proposes a novel approach to recover garment materials and patterns with optimization using differentiable simulation. While these solutions yield notable results, a gap persists in user-friendliness and practicality compared to direct communication of outcomes through natural language. Furthermore, prior studies have largely overlooked generating garment color, texture, and material, essential elements for creating high-quality garments.

\paragraph{Text-to-3D Generation}
Recent breakthroughs in the text-to-image domain \cite{ho2020denoising, zhang2023adding, rombach2022high} have enhanced interest in text-guided 3D content generation. Early work \cite{jain2022zero} introduces a text-to-3D method guided by CLIP \cite{radford2021learning}. \cite{poole2022dreamfusion, wang2023score} present the Score Distillation Sampling (SDS) algorithm, elevating pre-trained 2D diffusion models for the 3D generation. \cite{metzer2023latent} optimizes Neural Radiance Fields (NeRF) \cite{mildenhall2021nerf} in the latent space. \cite{lin2023magic3d, chen2023fantasia3d} optimize efficient mesh representations \cite{munkberg2022extracting, shen2021deep} for higher quality generation. \cite{seo2023let} integrates 3D awareness into 2D diffusion for improving text-to-3D generation consistency. Subsequent studies \cite{chen2023text2tex, lugmayr2022repaint, richardson2023texture} focus on texturing pre-existing meshes, balancing speed and quality. Despite their innovations, SDS-based methods faced challenges with over-saturation. \cite{tsalicoglou2023textmesh} proposes a novel method to refine mesh textures for more realistic generations. ProlificDreamer \cite{wang2023prolificdreamer} introduces the Variational Score Distillation (VSD) method to mitigate over-saturation effectively. Furthermore, several studies \cite{shi2023mvdream, ye2023consistent, zhao2023efficientdreamer, liu2023syncdreamer} explored 3D generation using multi-view diffusion. Concurrently, some research \cite{huang2023tech, liu2023one, liu2023zero, long2023wonder3d, qian2023magic123, raj2023dreambooth3d, shi2023zero123++, tang2023make, xu2023neurallift, wu2023hyperdreamer, melas2023realfusion} concentrates on reconstructing 3D content from a single image through distillation, achieving high-fidelity textured meshes from 2D diffusion priors. Additionally, some studies \cite{yu2023surf, erkocc2023hyperdiffusion, nash2020polygen, siddiqui2023meshgpt} delve into shape generation. \cite{yu2023surf} generates high-quality 3D shapes with Unsigned Distance Field (UDF) through Diffusion models. \cite{nash2020polygen, siddiqui2023meshgpt} employ autoregressive models for mesh structure generation. Although some general object generation methods \cite{qiu2023richdreamer, mildenhall2021nerf, poole2022dreamfusion, wang2023prolificdreamer, wang2023score, yu2023surf} can produce garments, their practicality in CG environments is limited. Very Recent work, Garment3DGen \cite{sarafianos2024garment3dgen} enables users to generate textured 3D garments from single images or text prompts based on a 3D base mesh. These 3D outputs, mostly mesh-based or derived from implicit fields, lack adaptability for fitting different bodies and layering multiple garments, common needs in garment design. Furthermore, the textures, typically produced via optimization or multi-view reconstruction, are often low-resolution and blurry, neglecting the structured UV mapping of garments, resulting in poor topology challenging for subsequent CG processing.

\section{Sewing Pattern Generation}
\label{sec:method1}

\begin{figure*} [ht]
  \centering
  \includegraphics[width=\textwidth]{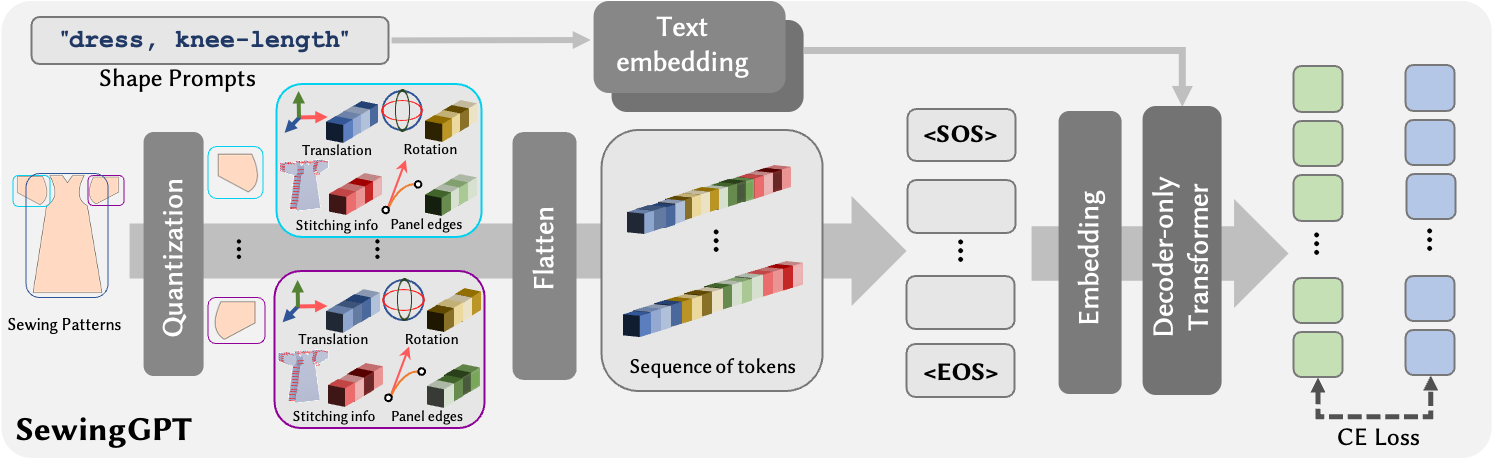}
  \captionof{figure}{\textbf{Overview of our SewingGPT pipeline.} We quantize sewing patterns to the sequence of tokens and adopt a GPT-based architecture to generate the tokens autoregressively. Our SewingGPT enables users to generate highly diverse and high-quality sewing patterns under text prompt guidance.} \label{fig_overview1}
\end{figure*}

\begin{figure}[t]
  \includegraphics[width=0.95\columnwidth]{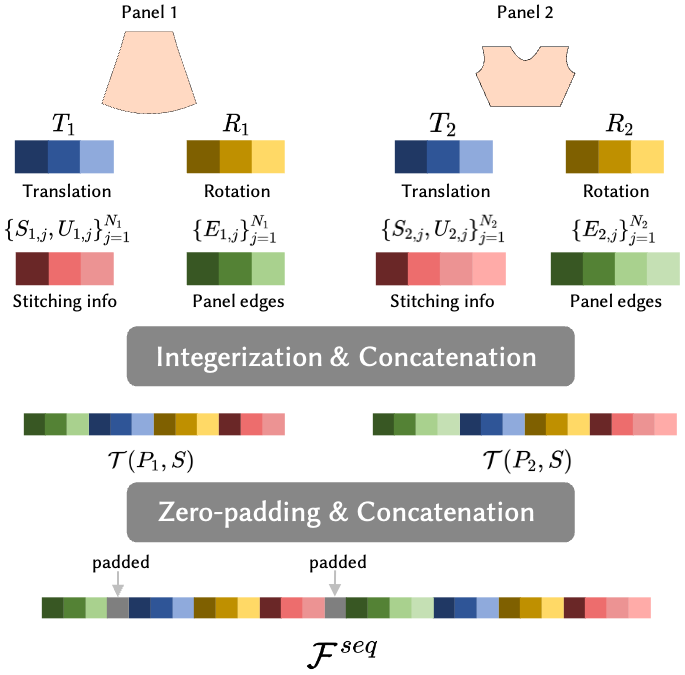}
  \captionof{figure}{\textbf{Details of our quantization.} We present an example of a part of a sleeveless dress, including a \textit{skirt panel} (Panel 1) and a \textit{top panel} (Panel 2). Assuming $N_1 < N_2 = K$, we require zero-padding for tokens from Panel 1.}
  \label{fig_pipeline3}
\end{figure}

Inspired by powerful language generative models, we introduce SewingGPT, a GPT-based autoregressive model for sewing pattern generation with text prompts. We first convert sewing pattern parameters into a sequence of quantized tokens and train a masked Transformer decoder, integrating cross-attention with text-conditioned embeddings. After training, our model can generate token sequences autoregressively based on user conditions. The generated sequences are then de-quantized to reconstruct the sewing patterns.

\subsection{Sewing Pattern Quantization}

\paragraph{Pattern representation.} We utilize the sewing pattern templates from \cite{korosteleva2022neuraltailor}, which cover a wide variety of garment shapes. Each sewing pattern includes $N_P$ panels $\{ P_i\}_{i = 1} ^ {N_P}$ and stitching information $S$. Each panel $P_i$ forms a closed 2D polygon with $N_i$ edges $\{ E_{i, j}\}_{j = 1}^{N_i}$. Each edge $E_{i, j}$ consists of four parameters $(v_x, v_y, c_x, c_y)$, where $(v_x, v_y)$ represents the edge's start point, and $(c_x, c_y)$ represents the control point of the Bezier curve. Since the panels form closed polygons, we do not need to store the edges' endpoints. The 3D placement of each panel is indicated by rotation quaternion $R_i \in \text{SO}(3)$ and translation vector $T_i \in \mathbb{R} ^ 3$. For stitching information, we utilize per-edge stitch tags $\{ S_{i, j}\}_{j = 1}^{N_i}$ and stitch flags $\{ U_{i, j}\}_{j = 1}^{N_i}$ for each panel $P_i$, obtained from the stitching information $S$. Each stitch tag $S_{i, j} \in \mathbb{R} ^ 3$ is based on the 3D placement of the corresponding edge, and each stitch flag is a binary flag with $U_{i, j} = \{0, 1\}$ indicating whether there is a stitch on this edge. We follow a similar approach to that used in \cite{korosteleva2022neuraltailor}, which utilizes Euclidean distance between tags as a similarity measure. To restore the stitching information from stitch tags and stitch flags, we filter out free and connected edges with stitch flags and then compare the stitch tags of all pairs of connected edges.

\paragraph{Quantization.} For each panel, we first utilize a similar data preprocessing approach to that used in \cite{korosteleva2022neuraltailor}, which standardizes all edge vectors and control points to maintain the data within a standard normal distribution, and normalizes its 3D placement to ensure all values are between 0 and 1. Then, we quantize all parameters, subsequently converting them into tokens. Specifically, for panel $P_i$, we model all parameters as discrete variables by multiplying predefined constants $C_E, C_R, C_T, C_S$ by edge vectors, rotation, translation, and stitching feature vectors respectively, and maintain stitching flags as 0 or 1. We flatten and concatenate $N_i$ edges, one rotation quaternion, one translation vector, $N_i$ stitching vectors, and $N_i$ stitching flags, into a sequence of tokens. We carefully select these constants to offer a good trade-off between maintaining the fidelity of sewing patterns and managing the vocabulary size. Overall, we can represent the quantization process as 
\begin{equation}
    \mathcal{T}(P_i, S) = C_E \{E_{i, j} \}_{j = 1} ^ {N_i} \oplus C_R  R_i \oplus C_T  T_i \oplus C_S  \{S_{i, j} \}_{j = 1} ^ {N_i} \oplus \{U_{i, j} \}_{j = 1} ^ {N_i}
\end{equation}
where we denote $\mathcal{T}$ as the quantization function, and $\oplus$ as the linear concatenation of tokens. These tokens are then formed into a linear sequence. We set a maximum limit, denoted as $K$, for the number of edges in each panel. To maintain a uniform token count across panels, we apply zero-padding to panels with $N_i < K$. Subsequently, all panels are flattened and merged into a single sequence, starting with a start token and ending with an end token. Owing to the uniformity of token counts for each panel, inserting padding tokens between panels is not required. Consequently, the resultant sequence, as illustrated in Figure~\ref{fig_pipeline3}, denoted as $\mathcal{F^{\text{seq}}}$, spans a length of $L_t = (8K + 7) N_P$, with each token denoted by $f_n$ for $n = 1, \ldots, L_t$. Eventually, we can represent it as
\begin{equation}
    \mathcal{F^{\text{seq}}} = \{ \mathcal{T} (P_i, S) + C\}_{i = 1} ^ {N_P},
\end{equation}
where C is a constant to ensure all tokens are non-negative.

\subsection{Generation with Autoregressive Model}
Utilizing GPT-based architectures, we adopt a decoder-only transformer to generate token sequences for sewing patterns. Inspired by PolyGen \cite{nash2020polygen}, we design the triple embedding for each input token: positional embedding, denoting which panel it belongs to; parameter embedding, classifying the token as edge coordinates, rotation, translation, or stitching feature vectors; and value embedding for the quantized sewing pattern values. The source tokens are then input into the transformer decoder to predict the probability distribution of the next potential token at each step. Consequently, the objective is to maximize the log-likelihood of the training sequences:
\begin{equation}
\mathcal{L} = -\prod_{i = 1}^{L_t} p\left(f_i \mid f_{< i}; \theta\right).
\end{equation}
By optimizing this objective, our SewingGPT learns the intricate relationships among the shape, placement, and stitching information of each panel. In the inference stage, target tokens begin with the start token and are recursively sampled from the predicted distribution $p\left(\hat{f}_i \mid \hat{f}_{< i}; \theta\right) $ until the end token. Following autoregressive token sequence generation, we reverse token quantization, converting the generated data to its original sewing pattern representation.

\paragraph{Conditional Generation with Text Prompts.} To guide the sewing pattern generation, our model integrates cross-attention with text-conditioned embeddings $\textbf{\text{h}}$. Initially, we utilize the CLIP model to obtain the CLIP embedding from input text prompts. Then, we project it into a feature embedding through a trainable compact Multilayer Perceptron (MLP) to condense the dimensionality of CLIP embeddings, matching the Transformer's dimensionality. This approach also boosts memory efficiency and inference speed. Subsequently, the Transformer decoder conducts cross-attention with the feature embedding \cite{li2022blip}. We train the model with pairwise data, facilitating condition-specific token generation.

\begin{figure} [t]
  \centering
  \includegraphics[width=\columnwidth]{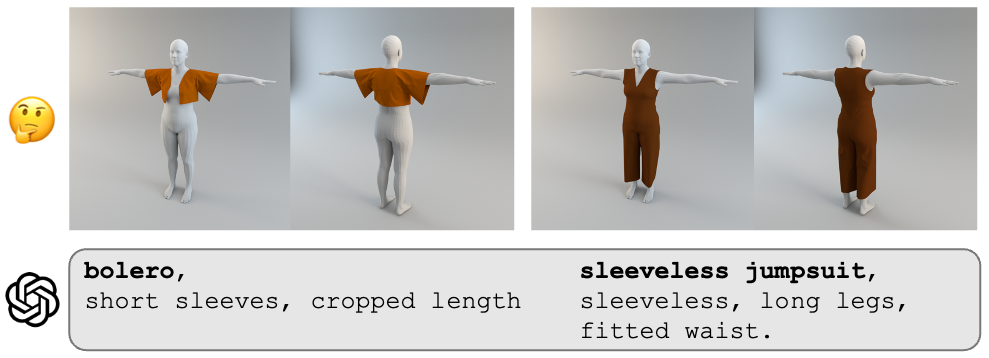}
  \captionof{figure}{\textbf{Examples of our data captions.} We utilize the rendered images and ask GPT-4V with the designed prompt for detailed captions.}
  \label{fig_data_caption}
\end{figure}

\begin{figure*} [t]
  \centering
  \includegraphics[width=\textwidth]{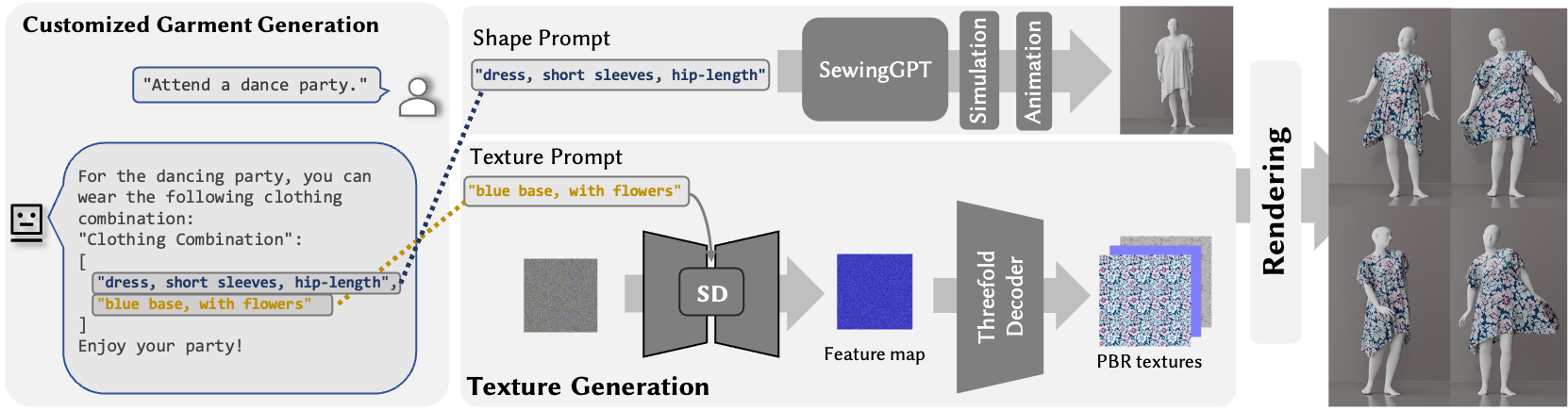}
  \captionof{figure}{\textbf{Overview of our entire DressCode pipeline for customized garment generation.} We employ a large language model to obtain shape prompts and texture prompts with natural language interaction and utilize the SewingGPT and a fine-tuned Stable Diffusion for high-quality and CG-friendly garment generation.} \label{fig_overview2}
\end{figure*}

\subsection{Implementation Details} 
    \paragraph{Dataset.} \label{data_processing}
        We utilize the extensive sewing pattern garment dataset from \cite{KorostelevaGarmentData}, notable for its comprehensive range of sewing patterns and styles of garments, including shirts, hoods, jackets, dresses, pants, skirts, jumpsuits, vests, etc. Our experiments use approximately 19264 samples across 11 fundamental categories. Each garment in the dataset contains a sewing pattern file, a 3D garment mesh draped on a T-pose human model, and a rendered image. We employ the GPT-4V \cite{achiam2023gpt} to generate captions for garments from the rendered images of the front view and the back view. For each garment, we first prompt GPT-4V to generate its common name (e.g., hood, T-shirt, blouse) if available, followed by a request for specific geometric features (e.g., long sleeves, wide garment, deep collar), as demonstrated in Figure~\ref{fig_data_caption}. We combine these two descriptions to form the caption for each garment. In our experiments, we use the pre-defined order of panels in the dataset for training. Additionally, We utilize about 90\% of the data from each category for training and the remaining for validation.
    \paragraph{Training.}
        We set $K = 14$, and the maximum length of tokens is $1500$. Our decoder-only Transformer consists of $24$ layers with position embedding dimensionality of $d_{pos} = 512$, parameter embedding dimensionality of $d_{para} = 512$, value embedding dimensionality of $d_{val} = 512$, and text feature embedding dimensionality of $d_f = 512$. We set constants $C_E = 50, C_R = 1000, C_T = 1000, C_S = 1000$ and, $C = 1000$. Our CLIP embeddings have a dimension of $d_{\text{CLIP}}=1024$, and condensed feature embeddings have a dimension of $d_{\text{feature}}=512$. We train our model using Adam optimizer, with a learning rate of $10 ^ {-4}$ and a batch size of $4$. The model is trained on a single A6000 GPU for 30 hours.

\section{Customized Garment Generation}
\label{sec:method2}
    
With SewingGPT, we have the capability to generate diverse sewing patterns directly from text prompts. Recognizing appearance's crucial role in the CG pipeline, we aim to generate corresponding Physically-based Rendering (PBR) textures for these patterns, aligning more closely with garment design workflows. By leveraging the SewingGPT and PBR texture generator, our framework DressCode further utilizes a large language model to create customized garments for users through natural language interaction.

\subsection{PBR Texture Generation}

In some production software commonly utilized by fashion designers, designers often create the texture of the garments after completing the pattern design. For garments, designers usually employ tile-based and physically-based textures such as diffuse, roughness, and normal maps to enhance the realistic appearance of the fabric. Therefore, to generate customized garments, we tailor a pre-trained Stable Diffusion \cite{rombach2022high} and employ a progressive training approach to generate PBR textures guided by text.

\paragraph{Latent Diffusion Finetuning.} 

Text-to-image generation has advanced significantly with the latent diffusion model (LDM). Existing foundation models, such as Stable Diffusion, trained on billions of data points, demonstrate extensive generalization capabilities. As the original LDM is trained on natural images, adapting it to generate tile-based images is necessary. To achieve this while maintaining the model's generalizability, we collect a PBR dataset with captions and fine-tune the pre-trained LDM on this dataset. We freeze the original encoder $\mathcal{E}$ and decoder $\mathcal{D}$, fine-tuning the U-Net denoiser at this stage. During inference, our fine-tuned LDM is capable of generating tile-based diffuse maps using text prompts.

\paragraph{VAE Finetuning.} As we can generate high-quality and tile-based diffuse maps, achieving realistic CG rendering requires us to further generate normal maps $U_n$ and roughness maps $U_r$ based on our generated diffuse maps $U_d$. In addition to the pretrained LDM encoder $\mathcal{E}$ and decoder $\mathcal{D}$, we fine-tune another two specific decoders $\mathcal{D}_n$ and $\mathcal{D}_r$. With a denoised texture latent code $z$ by text input, which can be decoded into diffuse maps through $D$, we utilize $\mathcal{D}_n$ and $\mathcal{D}_r$ to decode $z$ into normal maps and roughness maps respectively.

\subsection{Customized Generation through User-friendly Interaction}

\paragraph{Guided by natural language.} Following the implementation of generating sewing patterns and textures through text prompts, our framework in practical scenarios enables designers to interact with the generator using natural language instead of relying on dataset-like formatted prompts. We adopt GPT-4 \cite{achiam2023gpt} with content learning to interpret users' natural language inputs, subsequently producing shape prompts and texture prompts. These prompts are then fed to the SewingGPT and the PBR texture generator, respectively. Once sewing patterns are generated, we stitch them onto a T-pose human model. Subsequently, the generated garments, along with PBR textures, seamlessly integrate into industrial software, allowing for animation with the human model and rendering under various lighting, ensuring vivid, realistic results.

\begin{figure}[t]
  \includegraphics[width=0.97\columnwidth]{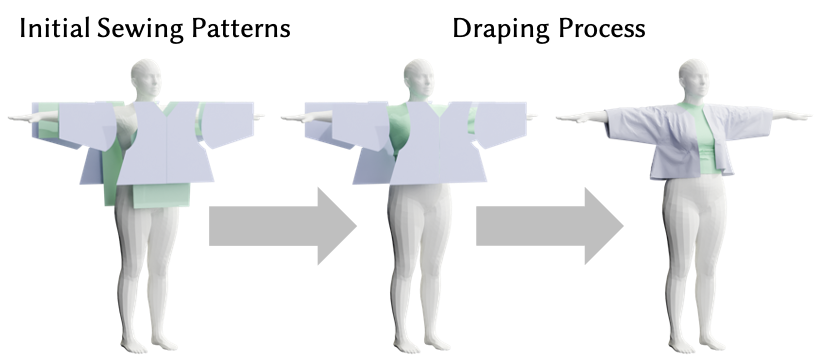}
  \captionof{figure}{\textbf{Examples of our multiple garments draping process.} Starting with initial sewing patterns, we first drape the inside T-shirt, followed by draping the outside jacket onto the model’s body.}
  \label{fig_draping}
\end{figure}

\paragraph{Multiple garments draping.} Production settings usually necessitate generating multiple clothing items (e.g., daily outfits like pants, T-shirts, and jackets) simultaneously. Past 3D content generation studies based on mesh or implicit fields face challenges in effectively achieving layered draping of multiple garments on a target human model. The adoption of sewing pattern representation enables the respective generation of multiple garments and their natural draping onto the human model. In our work, for T-pose results, we use the Qualoth simulator \cite{choi2002stable} as the physics simulator. We utilize the same material parameters and 3D human model from \cite{KorostelevaGarmentData}. In the process of draping multiple garments, we employ an automated sequential multi-garment draping technique, as depicted in Figure~\ref{fig_draping}. Specifically, for a set of clothes, we drape the garment onto the model’s body from the inside out. After each simulation, we combine the mesh of the simulated garment and the human model, then perform the next simulation with the subsequent garment.

\begin{figure}[t]
  \includegraphics[width=0.95\columnwidth]{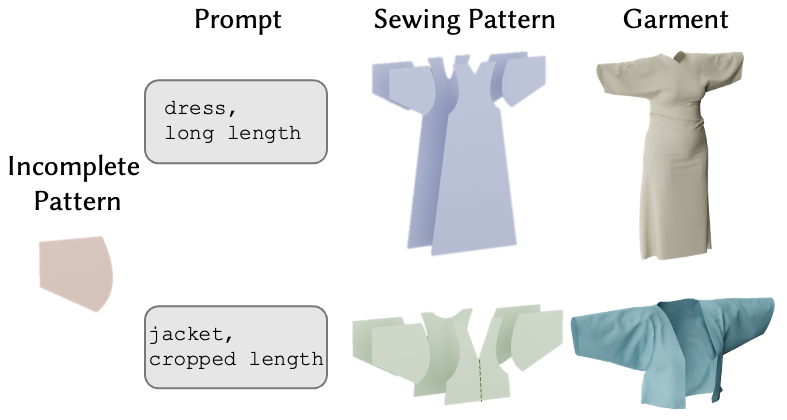}
  \captionof{figure}{\textbf{Examples of pattern completion.} Given an incomplete pattern, our method infers reasonable sewing pattern completions with various text prompts.} 
  \label{fig_shape_completion}
\end{figure}

\paragraph{Pattern Completion.}

Benefiting from the autoregressive model, our method can complete the entire sewing pattern by utilizing probabilistic predictions provided by the model upon receiving partial pattern information. Additionally, inputting a text prompt can guide the model in completing the sewing patterns. Our work, as illustrated in Figure~\ref{fig_shape_completion}, demonstrates that with a given sleeve, our model adapts to complete various sewing patterns based on different prompts. This enables users to design partial patterns manually and utilize SewingGPT for inspiration and completion of the garments guided by the text prompts.

\begin{figure}[t]
  \includegraphics[width=\columnwidth]{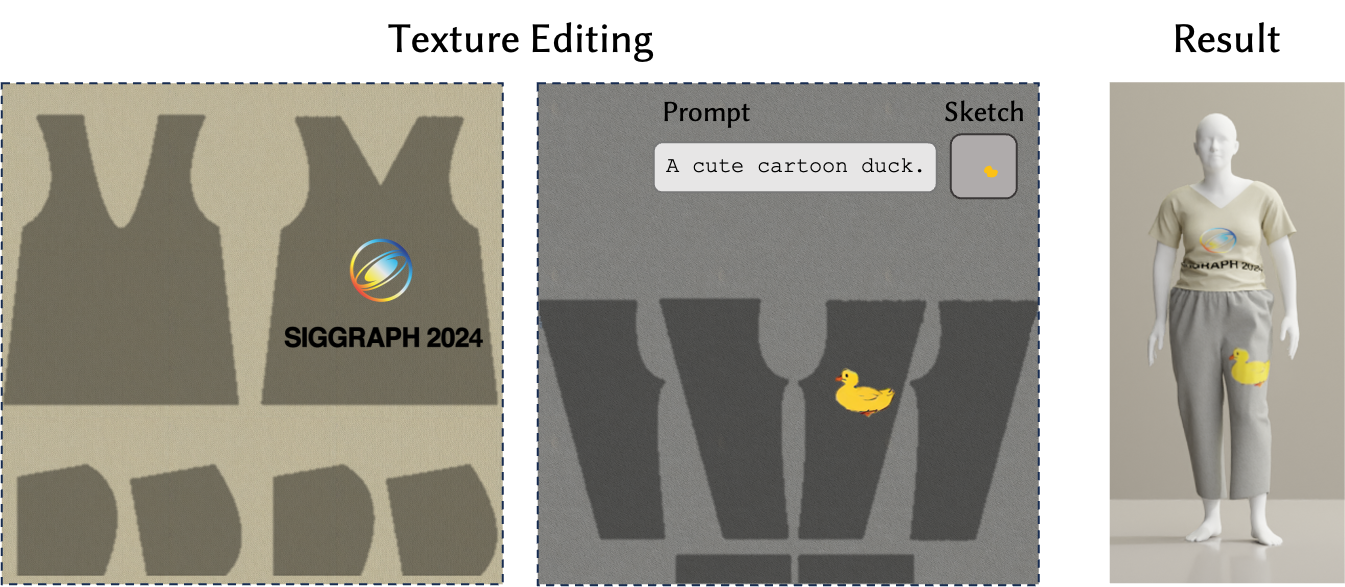}
  \captionof{figure}{\textbf{Examples of texture editing.} By manually drawing or creating sketches with text prompts, our method facilitates user-friendly texture editing.} 
  \label{fig_texture_editing}
\end{figure}

\begin{figure*} [h]
  \centering
  \includegraphics[height=0.91\textheight]{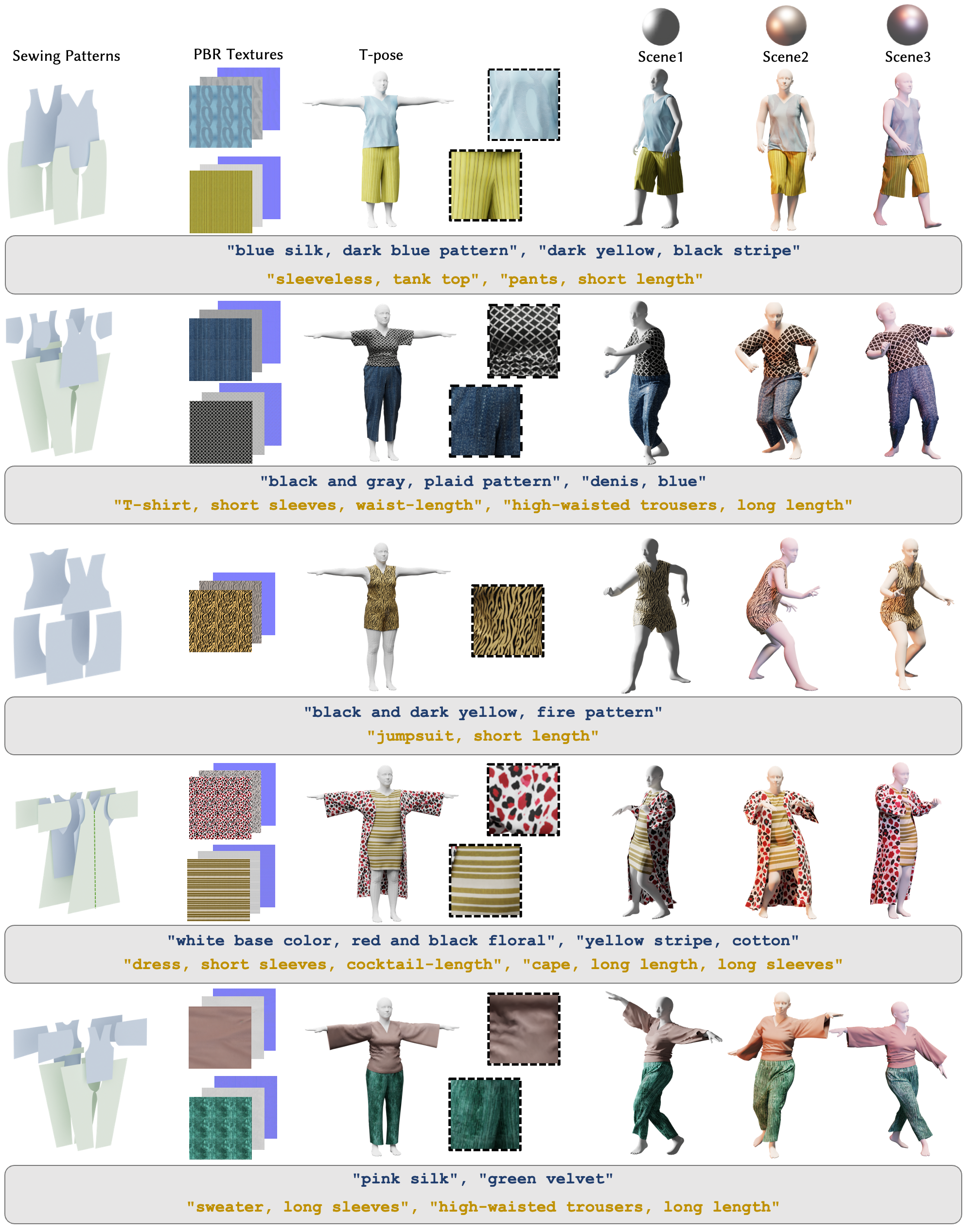}
  \captionof{figure}{\textbf{Our results gallery of DressCode.} We generate sewing patterns, PBR textures, and garments in diverse poses and lighting conditions, guided by various text prompts.} \label{fig_gallery}
\end{figure*}

\begin{figure*} [t]
  \centering
  \includegraphics[width=\textwidth]{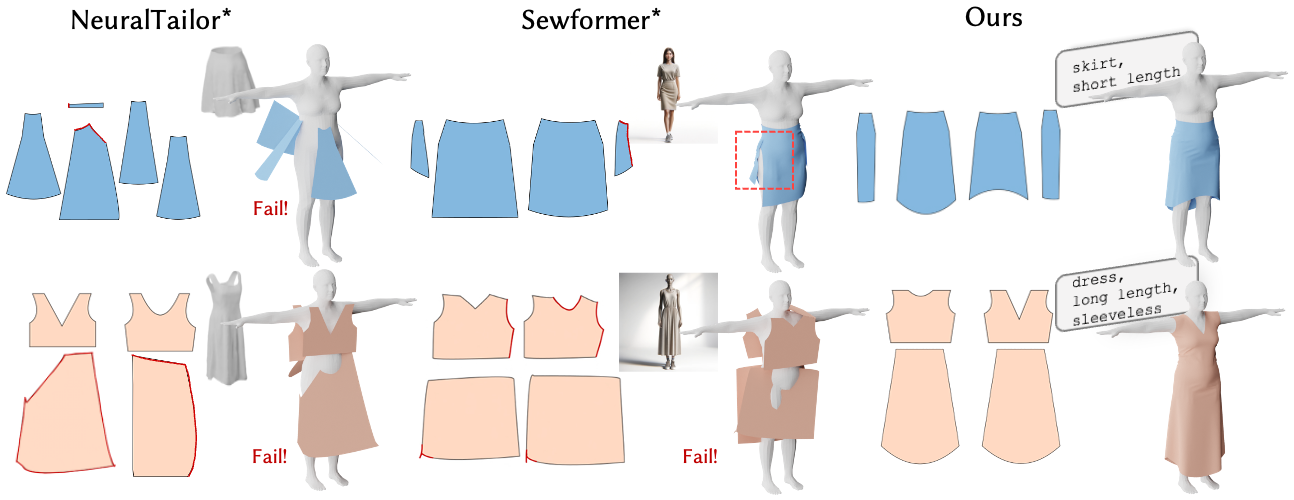}
  \captionof{figure}{\textbf{Qualitative comparisons on sewing patterns.} Major errors on panels are marked with red edges. The inputs of each method from left to right are meshes, images, and texts, respectively, which are shown along with the results.} \label{fig_comparison1}
\end{figure*}

\paragraph{Texture Editing.}

In the majority of recent 3D generation tasks, the inability to produce structured UV maps has been a significant impediment, particularly for generating garments. However, our generation method, utilizing sewing pattern representation, enables the creation of distinct and structured UV mappings of each panel. This facilitates convenient texture editing at specific locations, allowing efficient post-processing on the textures. As shown in  Figure~\ref{fig_texture_editing}, we demonstrate the SIGGRAPH icon drawn on a cream-colored T-shirt's diffuse map and a duck seamlessly blended with the original grey-colored pants' diffuse map by creating hand-made sketches with text prompts in the masked areas (refer to stable-diffusion-webui \cite{AUTOMATIC1111_Stable_Diffusion_Web_2022} for detailed implementation).

\section{Experiments}
\label{sec:experiments}

In this section, we first conduct qualitative and quantitative comparison experiments with other state-of-the-art 3D generation methods to demonstrate the generation capability of our method. We then present ablation studies and validation to evaluate our pipeline. Furthermore, we conduct a qualitative comparison experiment with parametric templates and perform a comprehensive user study to showcase our results compared to other methods. We also verify our results in Marvelous Designer \cite{marvelousdesigner}; we manually load generated garments into the software and then animate the garments with the human model in non-T-pose. We then show a results gallery in Figure~\ref{fig_gallery} of generated high-quality garments using our method with various text prompts, including generated sewing patterns, PBR textures, draping results on a T-pose human model, and animated results with various poses under different illuminations.

\begin{figure} [ht]
  \centering
  \includegraphics[width=\columnwidth]{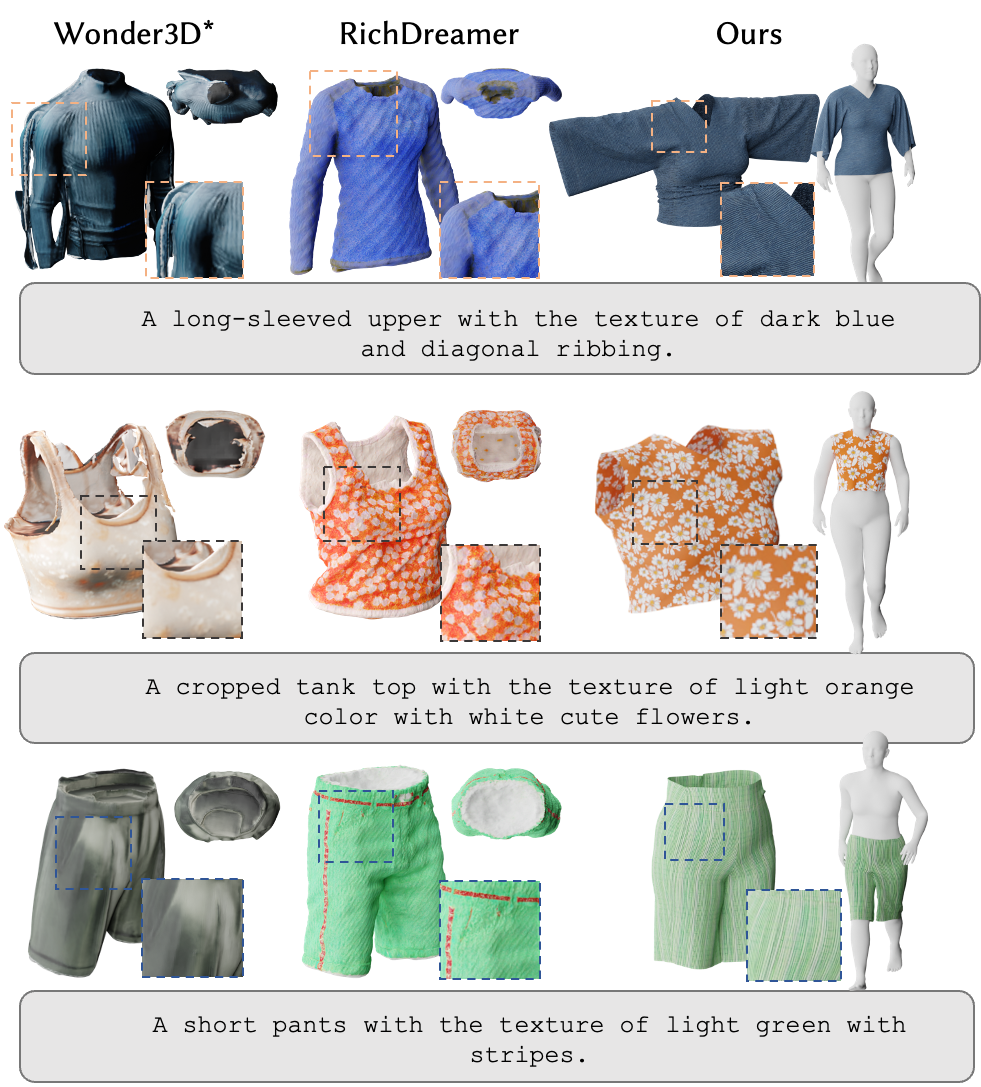}
  \captionof{figure}{\textbf{Qualitative comparisons on text-to-3D generation.} Images on the right side of our methods are the draped garments on human models. Images on the right top of other methods are the top view of each garment. We show our results yield high-quality rendering and the capability to drape on human bodies.} \label{fig_comparison2}
\end{figure}

\subsection{Comparison of 3D Garment Generation}

\paragraph{Comparisons on sewing pattern generation.}
We show some qualitative comparisons with two state-of-the-art sewing pattern works \cite{korosteleva2022neuraltailor, liu2023sewformer} in Figure~\ref{fig_comparison1}, where we present the panel prediction, draped garment on a T-pose human model, and corresponding inputs for each method. Given that NeuralTailor is designed for 3D point cloud inputs and trained on open-surface meshes, we utilize a 3D generation method Surf-D \cite{yu2023surf} using the UDF representation to create meshes conditioned on specific garment categories as inputs (denoted as \textbf{NeuralTailor*}). Note that Surf-D is trained on the Deep Fashion3D dataset. Although our method supports complex prompts, we only select category names such as \textit{skirt} and \textit{sleeveless dress}, which are presented in the Deep Fashion3D dataset, in our experiment as prompts to ensure a fair comparison. For Sewformer, designed for image inputs, we utilize DALLE-3 \cite{betker2023improving} to synthesize input images from text prompts (denoted as \textbf{Sewformer*}). While Sewformer is trained on images of human models wearing both upper and lower garments, we synthesize the images with the same rule and extract partial target panels from predicted results for comparison. For the first row of Figure~\ref{fig_comparison1}, our generated image involves a model wearing both a top shirt and a skirt, not only a sole skirt, which aligns with the training dataset in Sewformer. For fairness, we manually extract skirt patterns from Sewformer predictions, as shown in the comparison. Since the generated input meshes are mostly out of the domain of NeuralTailor's training dataset, the results appear as distorted panels and fail to be stitched together. Sewformer is trained on a new dataset with better generalization; nevertheless, it also encounters issues with irregular and distorted panels, as well as poor garments after stitching. Our method, yielding more accurate results, demonstrates robust generation capabilities with text prompts.

\begin{table}[t]
    \centering
    \begin{tabular}{c c c c}
    \toprule
    \quad & Wonder3D*  & RichDreamer  & Ours \\
    \midrule
    CLIP score $\uparrow$  &   0.302     &    0.324      &  \textbf{0.327} \\ 
    Runtime $\downarrow$  &  $\sim$ 4 mins      & $\sim$ 4 hours      & \textbf{$\sim$ 3 mins} \\ 
    PBR Texture         & \XSolidBrush   & \Checkmark   & \Checkmark \\
    Texture Editing          & \XSolidBrush & \XSolidBrush & \Checkmark \\
    Draping        & \XSolidBrush & \XSolidBrush & \Checkmark \\
    \bottomrule
    \end{tabular}
  \caption{\textbf{Quantitative and characteristic comparisons on different methods.} Compared to other methods, our method achieves the highest CLIP score and yields several CG-friendly characteristics.}
  \label{tab:method_comparison}
\end{table}

\paragraph{Comparisons on text-to-3D generation.}
We evaluate the quality of our customized garment generation with various 3D generation methods in Figure~\ref{fig_comparison2}. We present qualitative comparisons with two state-of-the-art 3D content generation methods: Wonder3D \cite{long2023wonder3d}, a 3D creation method from single images, and RichDreamer \cite{qiu2023richdreamer}, a text-to-3D work, generating with PBR textures. We adopt DALLE-3 \cite{betker2023improving} to synthesize image inputs for Wonder3D (denoted as \textbf{Wonder3D*}). Wonder3D takes about 4 minutes to generate garments but fails to retain fine detail and fidelity to the input images, yielding poor geometry. RichDreamer takes approximately 4 hours to optimize and yield more realistic results; however, the generated garments are still blurry for rendering. Furthermore, these generated garments are close-surface meshes, as shown in Figure~\ref{fig_comparison2}, and fail to adapt to human bodies. In contrast, our method takes about 1 minute to generate sewing patterns, and overall about 3 minutes to generate the simulated garments. It facilitates draping garments on human models in various poses and generating high-quality tile-based PBR textures, achieving realistic rendering.

Additionally, we adopt the CLIP score to quantitatively measure different methods. We generate 15 garments with highly diverse text prompts using each method. Then, we render the generated 3D garments with textures and calculate the CLIP score using the given text prompts. Although our method does not optimize the 3D models for fitting the rendered images to text prompts better, our model achieves the highest CLIP score, demonstrating the effectiveness of our method.
These general 3D methods are more broadly applicable than to only 3D garments. However, we also compare several characteristics among different methods, highlighting the advantages of our CG-friendly asset generation in the specific 3D garment domain. The results are shown in Table~\ref{tab:method_comparison}.

\begin{figure}[t]
  \includegraphics[width=\columnwidth]{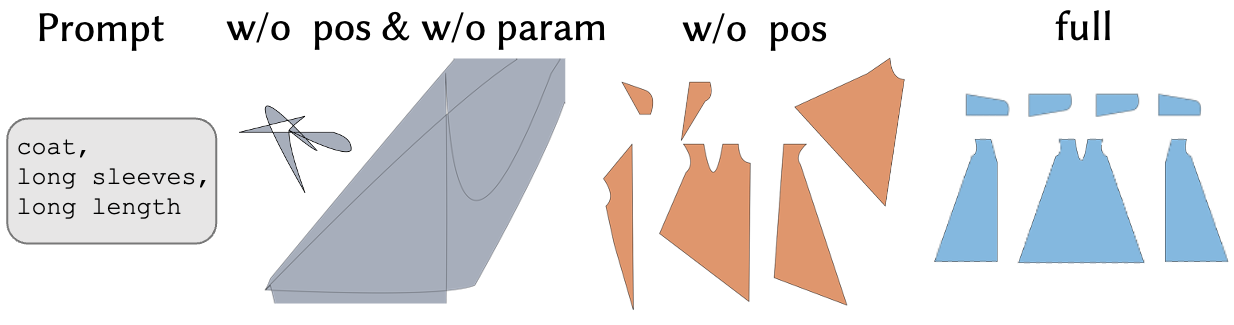}
  \captionof{figure}{\textbf{Ablation on embeddings.} Better sewing patterns are generated with our designed parameter embedding and positional embedding.} 
  \label{fig_ablation_study}
\end{figure}

\begin{figure}[t]
  \includegraphics[width=\columnwidth]{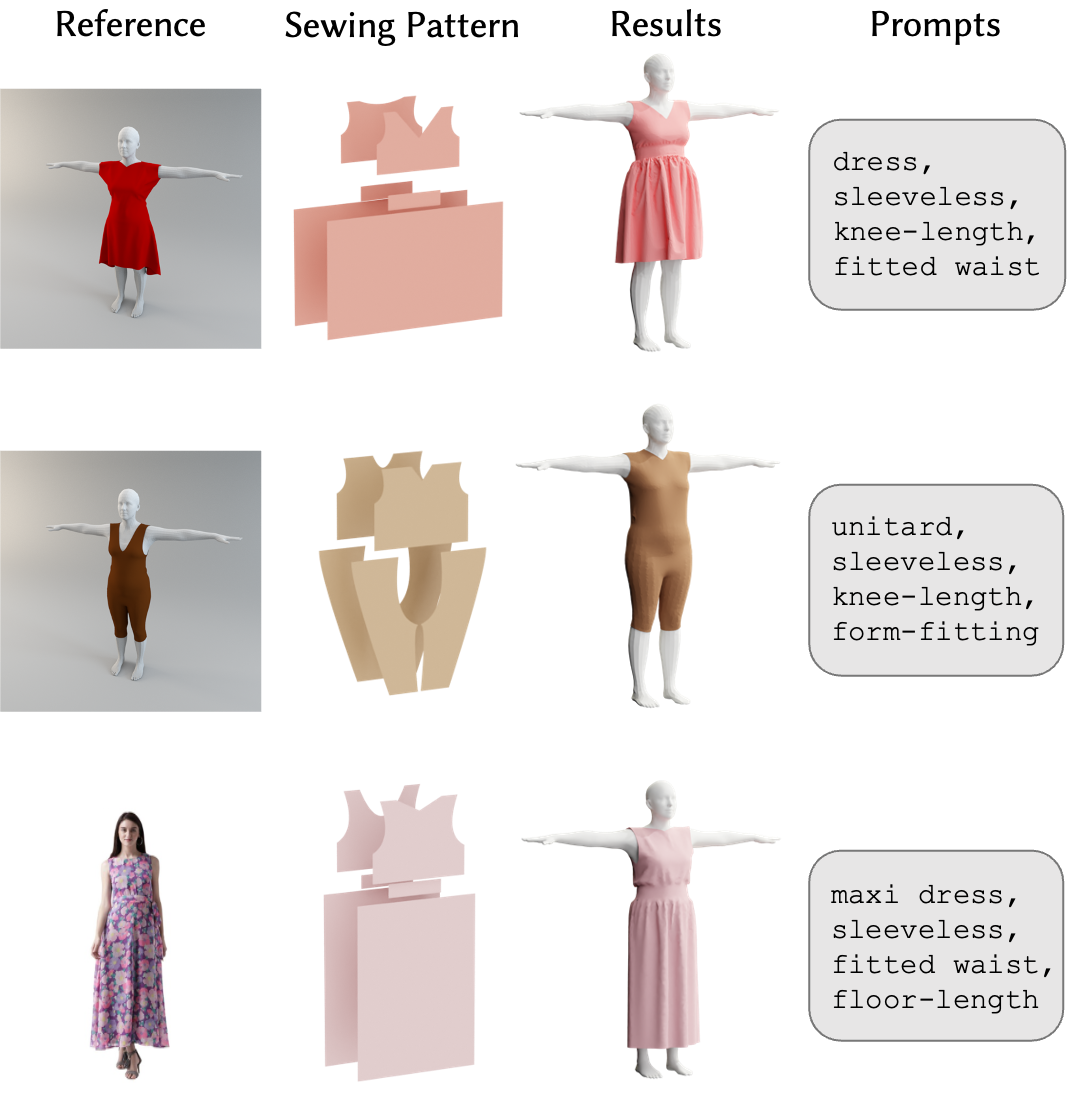}
  \captionof{figure}{\textbf{Validation.} We validate our generated garments with the text prompts from our dataset out of training and in the wild. Our results align well with the text prompts, demonstrating the effectiveness of our method.}
  \label{fig_validation}
\end{figure}

\subsection{Ablation Study}

We evaluate the performance of our triple embedding used in SewingGPT in Figure~\ref{fig_ablation_study}. We first train the model with only value embedding (denoted as \textbf{w/o pos \& w/o param}). The results are highly disordered, with strongly distorted panels, containing mismatched stitching information. We then incorporate the parameter embedding (denoted as \textbf{w/o pos}), facilitating the model to learn categories (e.g., edge coordinates, rotation parameters, translation parameters, or stitching feature vectors) of each token in a panel, and the results show the better shape of each panel, yet it is still distorted and lacking enough panels. Lastly, we further incorporate the positional embedding (denoted as \textbf{full}), enabling the model to distinguish between different panels and the number of panels, leading to the best and complete results.

\subsection{Validation}

To validate our methods, we design two experiments, as shown in Figure~\ref{fig_validation}. Firstly, we test text prompts from the holdout set as done for training and compare the results in the dataset with our generated outputs. In the first two rows of Figure~\ref{fig_validation}, we test two data points from outside the training dataset and observe that the generated results correspond well to the input prompts. Notably, in the dress example in the first row, although our generated result includes a waistband while the reference example does not, this still aligns well with the ``fitted waist'' attribute mentioned in the prompt. Furthermore, we test an image in the wild from the Deep Fashion3D dataset \cite{zhu2020deep} and utilize the method described in Section~\ref{data_processing} with GPT-4V \cite{achiam2023gpt} to generate the caption. This serves as the input to qualitatively compare our result with the reference image, revealing that our result closely resembles the input image. This demonstrates that our work effectively bridges the gap between conceptual textual descriptions and their practical, visual counterparts in garment design.

\begin{figure}
  \includegraphics[width=\columnwidth]{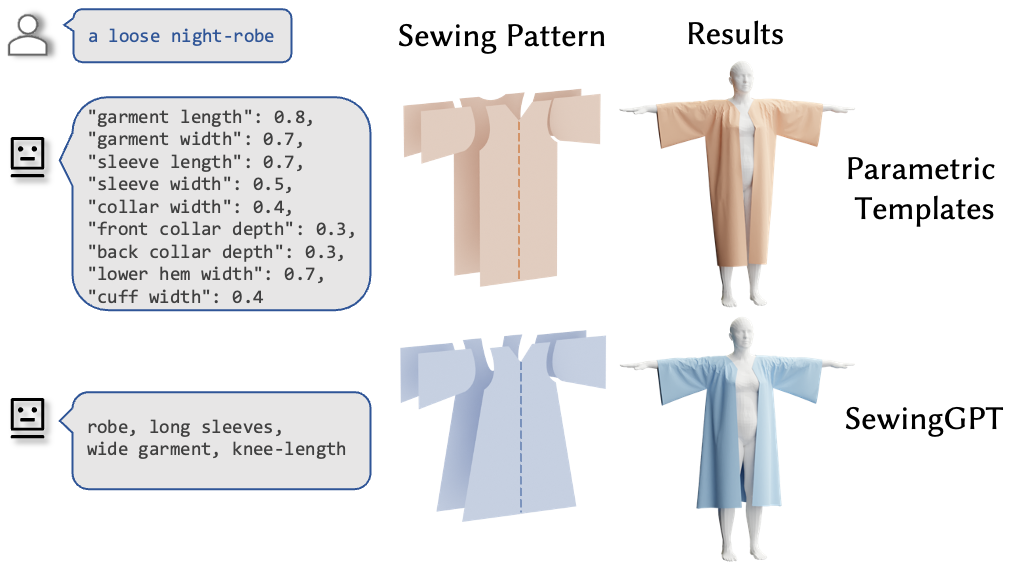}
  \captionof{figure}{\textbf{Qualitative comparison with Parametric Templates.} The user inputs the prompt ``a loose night-robe,'' and we show the results from Parametric Templates and our proposed SewingGPT.}
  \label{fig_gpt_formatted}
\end{figure}

\subsection{Comparison with Parametric Templates}

\cite{KorostelevaGarmentData} propose a method of a flexible description structure for specifying parametric sewing pattern templates. An intuitive idea is that we can control the parameters of the pre-defined parametric sewing pattern templates to generate diverse garments (denoted as \textbf{Parametric Templates}). To facilitate interaction through natural language and benefit from the strong capability of content learning in ChatGPT, we employ GPT-4 \cite{achiam2023gpt} in our experiment, designing prompts to enable its role as a garment design assistant, providing formatted outputs. Subsequently, we inquire about the description of garments, prompting it to output parameters for a specific template. As illustrated in the middle left of Figure~\ref{fig_gpt_formatted}, GPT-4 responds with several parameters when we ask with ``a loose night-robe.'' We qualitatively compare the results from Parametric Templates and our SewingGPT in Figure~\ref{fig_gpt_formatted}, observing that both methods generate reasonable results. Nevertheless, our proposed SewingGPT is more adaptable to the diverse categories of data, as it does not require selecting pre-defined templates and designing prompts specifically for ChatGPT's content learning. Additionally, our methods enable us to extend beyond such parametrized datasets for more complex sewing patterns.

\begin{figure}
  \includegraphics[width=\columnwidth]{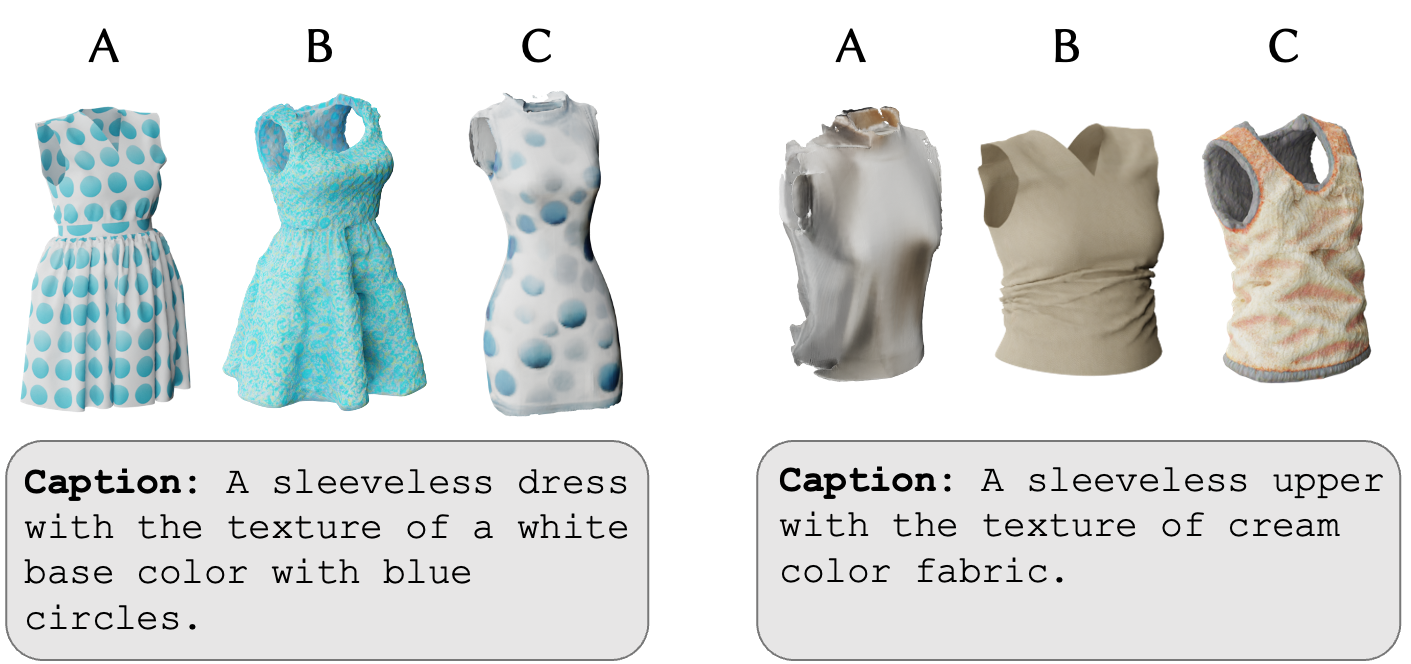}
  \caption{\textbf{Examples from our user study.} We present two cases for users to select the best results from three options: \textbf{A}, \textbf{B}, and \textbf{C}. The left group of results is generated by \textit{Ours}, \textit{RichDreamer}, and \textit{Wonder3D*}, respectively, and the right group of results is generated by \textit{Wonder3D*}, \textit{Ours}, and \textit{RichDreamer}, respectively. The generation method of each image is not disclosed to the users.}
  \label{fig_user_study_exp}
\end{figure}

\begin{table}
  \includegraphics[width=0.95\columnwidth]{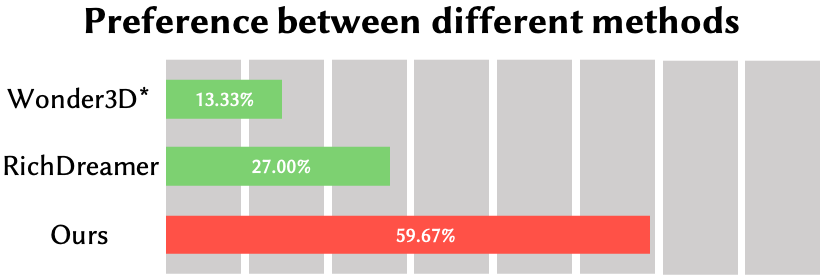}
  \caption{\textbf{Quantitative results of user study.} It demonstrates the higher user preference for our method compared to other methods.}
  \label{fig_user_study}
\end{table}

\subsection{User Study}

We conduct a comprehensive user study to evaluate the quality of our generated customized garments, particularly their alignment with given text prompts and overall quality. Given 20 text prompts, including descriptions of the shape and texture of garments, we then render the generated results for each method and shuffle the order of results obtained by different methods, as shown in Figure~\ref{fig_user_study_exp}. Then we ask 30 users to select the best results from all candidates with comprehensive consideration for two aspects: conformity to the captions of prompt texts regarding garments' shape and texture, and the visualized quality and fidelity of the rendered garments. As illustrated in Table~\ref{fig_user_study}, the preference results clearly indicate a significant advantage of our method over competing approaches in both aspects, highlighting its superiority in aligning closely with the textual prompts and producing visually appealing and high-fidelity garments.

\subsection{Limitations and Discussions}

\begin{figure}[t]
  \includegraphics[width=\columnwidth]{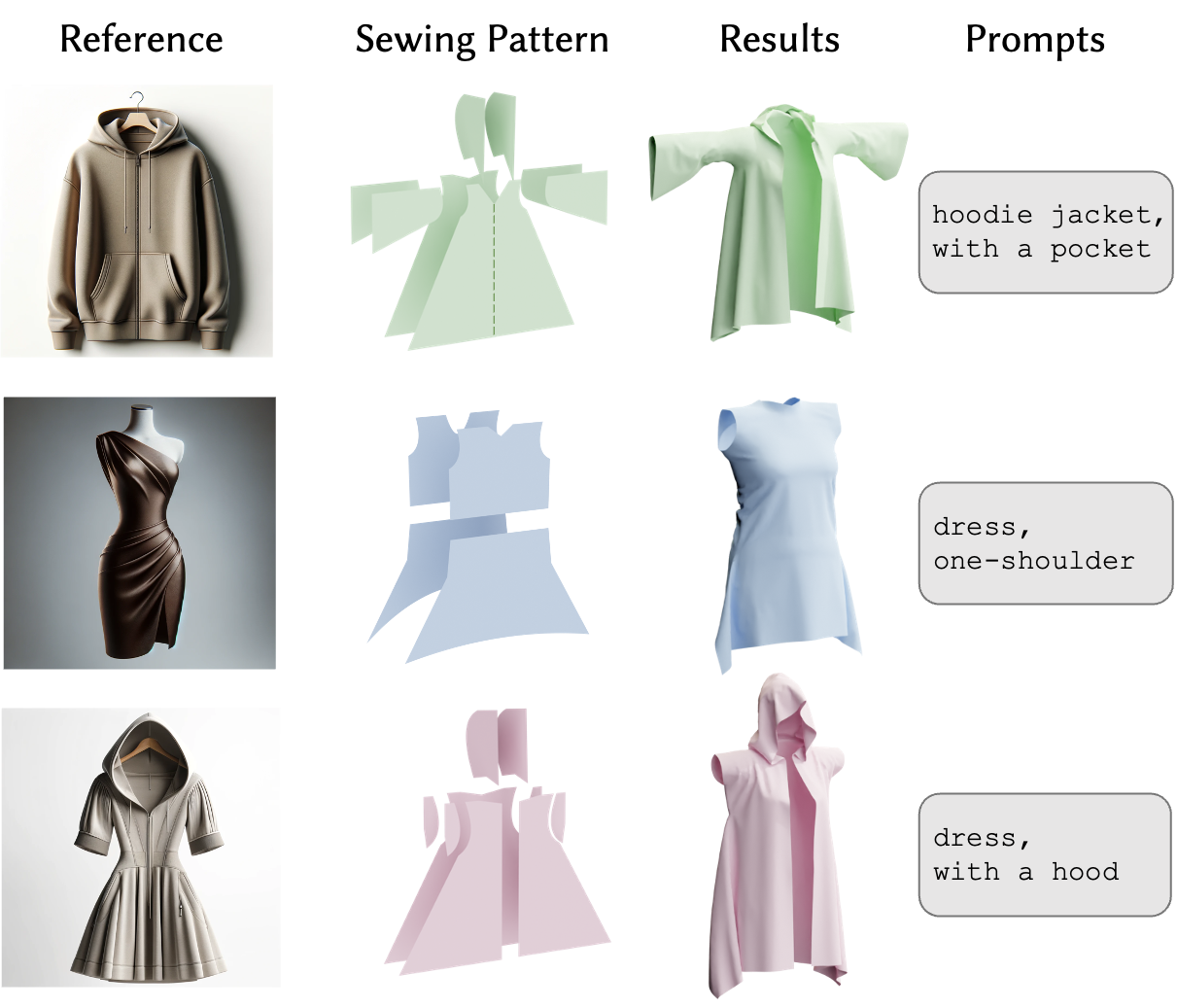}
  \caption{\textbf{Failure cases.} Our method struggles to generate garments that fall outside the domain of the training dataset. We present three examples of generated results: a ``hoodie jacket with a pocket,'' a ``one-shoulder dress,'' and a ``dress with a hood.'' The reference images are generated by DALLE-3 \cite{betker2023improving}.}
  \label{fig_limitation}
\end{figure}

Despite producing high-quality garment generation from text guidance, our method encounters certain limitations. One limitation is that the current sewing pattern dataset limits the generation of multi-layered garments, such as ``hoodie jacket with a pocket,'' as depicted in the first row of Figure~\ref{fig_limitation}. It underscores the importance of dataset expansion to include more complex stitching relationships. Another limitation is that our model struggles with prompts outside the domain of our dataset. For instance, we test prompts like ``one-shoulder dress.'' As shown in the second row of Figure~\ref{fig_limitation}, the model still generates a ``two-shoulder dress,'' due to the absence of ``one-shoulder'' garments in our dataset, which hinders its ability to recognize this attribute. Similarly, we experiment with integrating unusual characteristics into garments, combining specific attributes from different categories of garments, such as a ``dress with a hood,'' a style not commonly encountered in real life. Our results shown in the last row of Figure~\ref{fig_limitation} display a very loose hoodie jacket. Although the results somewhat resemble a dress with its loose style, a dress should not be open-front. This outcome is due to the presence of only hoodie jackets as hooded garments in our dataset, leading to a bias toward producing results within the hoodie jacket category when the prompt includes a ``hood.'' We believe that enriching the dataset with a wider variety of garments can significantly enhance the model's versatility. Additionally, inspired by recent breakthroughs in the generation domain, distilling knowledge from the pre-trained foundation model, such as SDS \cite{poole2022dreamfusion}, to improve generalization is a worthy direction for future work. Lastly, although our framework is pioneering in generating garments with text prompts, incorporating multi-modality inputs could prove more effective. Generating sewing patterns and textures controlled by both text and images presents a particularly intriguing and challenging problem yet to be addressed in real-world applications.

\paragraph{Potential ethical implications.} The text-driven generation method is subject to biases inherent in underlying pre-trained models such as CLIP and Stable Diffusion. Its user-friendliness and high-quality outputs also carry potential risks for misuse, emphasizing the necessity for future initiatives to tackle these ethical concerns through bias mitigation and thorough review. Additionally, the utilization of Stable Diffusion for both fine-tuning and inference purposes raises significant concerns regarding potential copyright issues, as the model may inadvertently generate content that mirrors proprietary works without explicit authorization. It is important for future work to ensure that the training data and generated content of these models are carefully reviewed and selected.

    
\section{Conclusions}
\label{sec:conclusions}
    In conclusion, this paper introduces \textit{DressCode}, a novel customized garment generation framework featuring a text-driven sewing pattern generator SewingGPT. This framework democratizes garment design by making it accessible and interactive, enabling both novices and experts to generate detailed sewing patterns and high-quality PBR textures through simple textual prompts. Additionally, our framework supports interaction-friendly applications for garment generation, completion, and editing, providing powerful tools that enable designers to unleash their creative imagination. Our experimental results and user study demonstrate the effectiveness of our method in producing CG-friendly garments that excel in quality and alignment with input prompts. 

In contrast to past work, our approach democratizes fashion design through enhanced accessibility and interactivity and improves the practical utility of digital garments in CG pipelines for post-editing, animation, and realistic rendering. The ease of use and innovative approach of DressCode promise exciting developments for future advancements in digital garments, potentially transforming digital garment creation and customization. We envision our method benefiting CG production and advancing the digital garment landscape for virtual try-on, fashion design, and digital human creation.


\begin{acks}
This work was supported by the National Key R\&D Program of China (2022YFF0902301), NSFC programs (61976138, 61977047), STCSM (2015F0203-000-06), and SHMEC(2019-01-07-00-01-E00003). We also acknowledge support from the Shanghai Frontiers Science Center of Human-centered Artificial Intelligence and MoE Key Lab of Intelligent Perception and Human-Machine Collaboration (ShanghaiTech University).
\end{acks}

\bibliographystyle{ACM-Reference-Format}
\bibliography{main}





\end{document}